\documentclass[11pt,a4paper]{article}
\usepackage[hyperref]{emnlp2018}
\usepackage{times}
\usepackage{latexsym}

\usepackage{url}

\aclfinalcopy % Uncomment this line for the final submission

\title{Team Papelo: Transformer Networks at FEVER}

\author{Christopher Malon \\
  NEC Laboratories America \\
  {\tt malon@nec-labs.com} \\}

\date{}

\begin{document}
\maketitle
\begin{abstract}
We develop a system for the FEVER fact extraction and verification
challenge that uses a high precision entailment classifier based on
transformer networks pretrained with language modeling, to classify a
broad set of potential evidence.  The precision of the entailment classifier
allows us to enhance recall by considering every statement from several
articles to decide upon each claim.  We include not only the articles best
matching the claim text by TFIDF score, but read additional articles whose
titles match named entities and capitalized expressions occurring in the
claim text.  The entailment module evaluates potential evidence one statement
at a time, together with the title of the page the evidence came from
(providing a hint about possible pronoun antecedents).
In preliminary evaluation, the
system achieves .5736 FEVER score, .6108 label accuracy, and .6485 evidence F1
on the FEVER shared task test set.
\end{abstract}

\section{Introduction}

The release of the FEVER fact extraction and verification dataset
\citep{thorne} provides a large-scale challenge that tests a combination of
retrieval and textual entailment capabilities.  To verify a claim in the
dataset as supported, refuted, or undecided, a system must retrieve
relevant articles and sentences from Wikipedia.  Then it must
decide whether each of those sentences, or some combination of them,
entails or refutes the claim, which is an entailment problem.  Systems
are evaluated on the accuracy of the claim predictions, with credit only given
when correct evidence is submitted.

As entailment data, premises in FEVER data differ substantially from those in
the image caption data used as the basis for the Stanford Natural Language
Inference (SNLI) \citep{snli} dataset.  Sentences are longer
(31 compared to 14 words on average), vocabulary is more abstract,
and the prevalence of named entities and out-of-vocabulary terms is higher.

The retrieval aspect of FEVER is not straightforward either.
A claim may have small word overlap with the relevant evidence,
especially if the claim is refuted by the evidence.

Our approach to FEVER is to fix the most obvious shortcomings
of the baseline approaches to retrieval and entailment, and to
train a sharp entailment classifier that can be used to filter
a broad set of retrieved potential evidence.
For the entailment classifier we compare
Decomposable Attention \citep{parikh, allennlp} as implemented in
the official baseline, ESIM \citep{esim},
and a transformer network with pre-trained weights \citep{openai}.
The transformer network naturally supports out-of-vocabulary
words and gives substantially higher performance than the other methods.

% Releasing code - at least entailment transformer
% Releasing fever-title-one
% Releasing fever-baselines modifications or code to set up the training

% Recently, OpenAI has released \citep{openai} a transformer network
% with pre-trained weights.  

\section{Transformer network}

The core of our system is an entailment module based on a transformer network.
Transformer networks \citep{vaswani} are deep networks applied to sequential
input data, with each layer implementing multiple heads of scaled dot product
attention.  This attention mechanism allows deep features to be
compared across positions in the input.

Many entailment networks have two sequence inputs, but the transformer
is designed with just one.
A separator token divides the premise from the hypothesis.

We use a specific transformer network released by OpenAI \citep{openai}
that has been pre-trained for language
modeling.\footnote{https://github.com/cdmalon/finetune-transformer-lm}
The network consists of twelve blocks.
Each block consists of a multi-head masked self-attention layer,
layer normalization \citep{layernorm}, a feed forward network,
and another layer normalization.  After the twelfth block, two
branches exist.  In one branch,
matrix multiplication and softmax layers are applied at the terminal
sequence position to predict the entailment classification.
In the other branch, a hidden state is multiplied by each token embedding
and a softmax is taken to predict the next token.
The language modeling branch has been pre-trained on the BookCorpus dataset
\citep{bookscorpus}.  We take the pre-trained model and train both
branches on examples from FEVER.

\section{\label{sec:entailment}Reframing entailment}

The baseline FEVER system \citep{thorne} ran the AllenNLP \citep{allennlp}
implementation of Decomposable Attention \citep{parikh}
to classify a group of five premise statements
concatenated together against the claim.  These five premise statements
were fixed by the retrieval module and not considered individually.
In our system, premise statements are individually evaluated.

\begin{table*}[t!]
\begin{center}
\begin{tabular}{lcccc}
\hline
Problem & \multicolumn{2}{c}{Support} & \multicolumn{2}{c}{Claim} \\
& Accuracy & Kappa & Accuracy & Kappa \\
\hline
ESIM on FEVER One & .760 & .260 & .517 & .297 \\
ESIM on FEVER Title One & .846 & .394 & .639 & .433 \\
Transformer on FEVER Title One & .958 & .660 & .823 & .622 \\
\hline
\end{tabular}
\caption{\label{table:title} Effect of adding titles to premises.}
\end{center}
\end{table*}

\begin{table*}[t!]
\begin{center}
\begin{tabular}{lcccc}
\hline
Problem & \multicolumn{2}{c}{Support} & \multicolumn{2}{c}{Claim} \\
& Accuracy & Kappa & Accuracy & Kappa \\
\hline
ESIM on FEVER Title Five Oracle & N/A & N/A & .591 & .388 \\
ESIM on FEVER Title Five & N/A & N/A & .573 & .110 \\
ESIM on FEVER Title One & .846 & .394 & .639 & .433 \\
Transformer on FEVER Title Five Oracle & N/A & N/A & .673 & .511 \\
Transformer on FEVER Title Five & N/A & N/A & .801 & .609 \\
Transformer on FEVER Title One & .958 & .660 & .823 & .622 \\
\hline
\end{tabular}
\caption{\label{table:five} Concatenating evidence or not.}
\end{center}
\end{table*}

We collect training data as the five sentences with the highest TFIDF
score against the claim, taken from the Wikipedia pages
selected by the retrieval module.  If any ground truth evidence group
for a claim requires more than one sentence, the claim is dropped from
the training set.  Otherwise, each sentence is labeled with the
truth value of the claim if it is in the ground truth evidence set,
and labeled as neutral if not.  The resulting data forms an entailment
problem that we call ``FEVER One.''
For comparison, we form ``FEVER Five'' and ``FEVER Five Oracle''
by concatenating all five retrieved sentences, as in the baseline.
In FEVER Five Oracle, the
ground truth is the claim ground truth (if verifiable), but in FEVER Five,
ground truth depends on whether the retrieved evidence is in the ground truth
evidence set.

Several FEVER claims require multiple statements as evidence in order to
be supported or refuted.
The number of such claims is relatively small: in the first
half of the development set, only 623 of 9999 claims were verifiable and had
no singleton evidence groups.  Furthermore,
we disagreed with many of these annotations and thought that
less evidence should have sufficed.  Thus we chose not to develop a strategy
for multiple evidence statements.

To compare results on FEVER Five to FEVER One, we must aggregate decisions
about individual sentences of possible evidence to a decision about the
claim.  We do this by applying the following rules:
\begin{enumerate}
\item If any piece of evidence supports the claim, we classify
the claim as supported.
\item If any piece of evidence refutes the claim, but no piece of evidence
supports it, we classify the claim as refuted.
\item If no piece of evidence supports or refutes the claim, we classify the
claim as not having enough information.
\end{enumerate}
We resolve conflicts between supporting and refuting information in
favor of the supporting information, because we observed cases in the
development data where information was retrieved for different entities
with the same name.  For example, Ann Richards appeared both as a
governor of Texas and as an Australian actress.  Information that would
be a contradiction regarding the actress should not stop evidence that would
support a claim about the politician.

Even if a sentence is in the evidence set, it might not be possible
for the classifier to correctly determine whether it supports the
claim, because the sentence could have pronouns with antecedents
outside the given sentence.  Ideally, a coreference resolution system could
add this information to the sentence, but running one could be
time consuming and introduce its own errors.  As a cheap alternative,
we make the classifier aware of the title of the Wikipedia page.
We convert any undersores in the page title to spaces, and insert the
title between brackets before the rest of each premise sentence.
The dataset constructed in this way is called ``FEVER Title One.''

\begin{table*}[t!]
\begin{center}
\begin{tabular}{lc}
\hline
System & Retrieval \\
\hline
FEVER Baseline (TFIDF) & 66.1\% \\
+ Titles in TFIDF & 68.3\% \\
+ Titles + NE & 80.8\% \\
+ Titles + NE + Film & 81.2\% \\
Entire Articles + NE + Film & 90.1\% \\
% /zdata/users/malon/fever/dev-tokenized.jsonl .6606  % no titles - original
% data-r49/dev-retrieved.jsonl .6833  % already includes titles
% data/no-film/dev-retrieved.jsonl .8083   % NER
% data/dev-retrieved.jsonl .8117   % NER, film
% dev-super-retrieved.jsonl .9009  % NER, film, whole articles
\hline
\end{tabular}
\caption{\label{table:retrieval} Percentage of evidence retrieved from first half of development set.  Single-evidence claims only.}
\end{center}
\end{table*}

\begin{table*}[t!]
\begin{center}
\begin{tabular}{lcc}
\hline
System & Development & Test \\
\hline
FEVER Title Five Oracle & .5289 & --- \\
FEVER Title Five & .5553 & --- \\
FEVER Title One & .5617 & .5539 \\
FEVER Title One (Narrow Evidence) & .5550 & --- \\
FEVER Title One (Entire Articles) & .5844 & .5736 \\
% Five-oracle (NER, Film) & .5289
% Five-self (NER, Film) % .5553
% NER, Film & .5617 & .5539
% NER, Film, only support & .5550 & ---    % don't send statements not predicted
% Super  & .5844 & .5736
\hline
\end{tabular}
\caption{\label{table:fever} FEVER Score of various systems.  All use NE+Film retrieval.}
\end{center}
\end{table*}

The FEVER baseline system works by solving FEVER Five Oracle.
Using Decomposable Attention, it achieves .505
accuracy on the test half of the development set.
Swapping in the Enhanced Sequential Inference
Model (ESIM) \citep{esim} to solve FEVER Five Oracle results in
an accuracy of .561.
Because ESIM uses a single out-of-vocabulary (OOV) token for all unknown
words, we expect it to confuse named entities.  Thus we extend the model
by allocating 10,000 indices for out-of-vocabulary words with randomly
initialized embeddings, and taking a hash
of each OOV word to select one of these indices.  With extended ESIM,
the accuracy is .586.
Therefore, we run most later comparisons with extended ESIM
or transformer networks
as the entailment module, rather than Decomposable Attention.
% This paragraph will be replaced with a comparison of all three networks
% on esim-title-one.

The FEVER One dataset is highly unbalanced in favor of neutral statements,
so that the majority class baseline would achieve 93.0\% on this data.
In fact it makes training ESIM a challenge, as the model only learns the
trivial majority class predictor if the natural training distribution is
followed.  We reweight the examples in FEVER One for ESIM so that each
class contributes to the loss equally.  Then, we use Cohen's Kappa rather
than the accuracy to evaluate a model's quality, so that following the bias
with purely random agreement is not rewarded in the evaluation.
In Table \ref{table:title}
we compare FEVER One to FEVER Title One, both at the level of classifying
individual support statements and of classifying the claim by aggregating
these decisions as described above.
On a support basis, we find a 52\% increase in Kappa by adding the titles.

When ESIM is replaced by the transformer network, class reweighting
is not necessary.  The network naturally learns to perform in excess
of the majority class baseline.  Cohen's Kappa is 68\% higher than
that for ESIM.  % move transformer FEVER Title One from in table to inline here?
% We suspect that ... already said stuff about BPE

The possibility of training on oracle labels for a concatenated set of
evidence allows a classifier to
simply guess whether the hypothesis is true and supported somewhere,
rather than having to consider the relationship between hypothesis
and premise.  For example, it is possible to classify 67\% of SNLI
examples correctly without reading the premise \citep{gururangan}.
As we show in Table \ref{table:five}, for ESIM, we find that
this kind of guessing makes the FEVER Title Five Oracle performance better
than FEVER Title Five.  The Transformer model is
accurate enough that oracle guessing does not help.  Both models perform
best when classifying each bit of evidence separately and then aggregating.

% Table A: formulations with ESIM (and maybe transformer)
% show that title helps classification
%   <-- does Juho have transformer with just fever-one no title?
%    otherwise I can show esim-title-one versus esim-one
%         (not true for five-oracle)

% Table B: differences between models, with title-one, ideally
% show the difference between decomposable, esim, transformer
%      For their decomposable baseline:
%        Use eval_da.py  with something on one basis   [No titles possible though]
%        Have json-confusion.pl evaluating their "five" version
%      To really train decomposable:
%           use oov supporting version
%      To train ESIM, with and without OOV

\section{Improving retrieval}

Regardless of how strong the entailment classifier is, FEVER score is
limited by whether the document and sentence retrieval modules, which produce
the input to the entailment classifier, find the right evidence.
In Table \ref{table:retrieval}, we examine the percentage of claims
for which correct evidence is retrieved, before filtering with the
entailment classifier.  For this calculation, we skip
any claim with an evidence group with multiple statements, and count
a claim as succesfully retrieved if it is not verifiable or if the statement
in one of the evidence groups is retrieved.
The baseline system
retrieves the five articles with the highest TFIDF score, and then extracts
the five
sentences from that collection with the highest TFIDF score against the claim.
It achieves 66.1\% evidence retrieval.

Our first modification simply adds the title to each premise statement when
computing its TFIDF against the claim, so that statements from a relevant
article get credit even if the subject is not repeated.  This raises
evidence retrieval to 68.3\%.

A more significant boost comes from retrieving additional Wikipedia pages based
on named entity recognition (NER).  We start with phrases tagged as
named entities by SpaCy \citep{spacy}, but these tags
are not very reliable, so we include various
capitalized phrases.  We retrieve Wikipedia pages whose title exactly matches
one of these phrases.

The named entity retrieval strategy boosts the evidence retrieval rate
to 80.8\%, while less than doubling the processing time.  However, sometimes
the named entity page thus retrieved is only a Wikipedia disambiguation page
with no useful information.  Noticing a lot of questions about films in the
development set, we modify the strategy to also retrieve a page titled
``X (film)'' if it exists, whenever ``X'' is retrieved.
The film retrievals raise evidence retrieval to 81.2\%.

Finally, we eliminate the TFIDF sentence ranking to expand sentence
retrieval from five sentences to entire articles, up to the first fifty
sentences from each.  Thus we obtain 2.6 million statements to classify
regarding the 19,998 claims in the shared task development set, for an
average of 128 premises per claim.
The evidence retrieval rate, including all these premises, increases to 90.1\%.
We continue to apply the entailment
module trained with only five premise retrievals.  Running the
entailment module on this batch using a machine with three NVIDIA GeForce
GTX 1080Ti GPU cards takes on the order of six hours.

Retrieving more than five sentences means that we can no longer submit
all retrieved evidence as support for the claims.  Instead, we follow the
aggregation strategy from Section \ref{sec:entailment} to decide the claim
label, and only submit statements whose classification matches.
Limiting evidence in this way when only five statements are retrieved
(``narrow evidence'' in Table \ref{table:fever})
pushes FEVER score down very little, to .5550 from .5617 on the development
set, so we have confidence that the extra retrieval will make up for the loss.
Indeed, when the system reviews the extra evidence, FEVER score goes up
to .5844 on the development set.

Table \ref{table:fever} compares the end-to-end performance of systems that
evaluate five retrieved statements together, evaluate five retrieved statements
separately, and evaluate all statements from entire articles separately.
Evaluating the statements separately gives better performance.
We submit the systems that retrieve five statements and entire articles
for evaluation on the test set, achieving preliminary FEVER scores
of .5539 and .5736 respectively (label accuracy of .5754 and .6108,
evidence recall of .6245 and .5002,
evidence F1 of .2542 and .6485).  In preliminary standings, the latter
system ranks fourth in FEVER score and first in evidence F1.

\section{Discussion}

Our approach to FEVER involves a minimum of heuristics and relies mainly
on the strength of the Transformer Network based entailment classification.
The main
performance gains come from adding retrievals that resolve named entities
rather than matching the claim text only, filtering fewer of the retrievals,
and making the entailment classifier somewhat aware of the topic of what
it is reading by including the title.  If higher quality and more plentiful
multi-evidence claims would be constructed, it would be nice to
incorporate dynamic retrievals into the system, allowing the classifier
to decide that it needs more information about keywords it encountered
during reading.

\bibliography{fever}
\bibliographystyle{aclnatbibnourl}

\end{document}